\documentclass[11pt,a4paper]{article}
\usepackage[hyperref]{eacl2021}
\usepackage{times}
\usepackage{latexsym}
\usepackage{float}

\usepackage{graphicx}

\usepackage{microtype}

\aclfinalcopy 

\title{Multi-Stream Transformers}

\author{Mikhail Burtsev \\
  Artificial Intelligence Research Institute \\
  Moscow, Russia \\
  Moscow Institute of Physics and Technology \\
  Dolgoprudny, Russia \\
  \texttt{burtsev@airi.net} \\\And
  Anna Rumshisky \\
   Univ. of Massachusetts Lowell,\\
   Lowell MA\\
  \texttt{arum@cs.uml.edu} \\}

\date{}

\begin{document}
\maketitle

\begin{abstract}

Transformer-based encoder-decoder models produce a fused token-wise representation after every encoder layer.  We investigate the effects of allowing the encoder to preserve and explore alternative hypotheses, combined at the end of the encoding process. To that end, we design and examine a \textit{Multi-stream Transformer} architecture and find that splitting the Transformer encoder into multiple encoder streams and allowing the model to merge multiple representational hypotheses improves performance, with further improvement obtained by adding a 
skip connection between the first and the final encoder layer.

\end{abstract}

\section{Introduction}

Over the past couple of years, Transformer-based architectures \cite{vaswani2017attention} have consistently shown state-of-the-art performance on a variety of natural language processing (NLP) benchmarks.
Transformer models implement an encoder-decoder architecture, in which encoder and decoder consist of several Transformer layers.  

Each Transformer layer is comprised by a self-attention module followed by a position-wise feed-forward layer.  
%
Self-attention computes three representations for each input token (query, key, and value), then updates the embedding for each token by computing a weighted sum of "value" representations for the other tokens.  The weights are given by a normalized dot product of the "query" representation for the original token with the "key" representations for the other tokens.
Skip connections are used around both the attention layer and the feed-forward layer, with layer normalization performed after each of them.
Token-wise representation computed by each Transformer layer allows for easier parallelization and therefore faster training.

While it is not entirely clear what serves as the main source of Transformer's
competitive advantage over other architectures, there have been claims that multi-layer Transformers pre-trained with a language modeling objective recover the traditional NLP pipeline \cite{tenney2019bert}. In particular, multiple probing studies found that earlier layers produce a better representation for "low-level" NLP tasks such as POS-tagging, while the representations computed at middle layers are better suited for recovering syntactic information \cite{tenney2019bert,hewitt2019structural,goldberg2019assessing,jawahar2019does,VigBelinkov_2019_Analyzing_Structure_of_Attention_in_Transformer_Language_Model}.


In the traditional multi-stage NLP pipeline architectures, every processing step usually ends with selection from multiple alternative hypotheses. For example, reading comprehension might depend on the results of coreference resolution, which in turn  depends on extracted mentions. Here different groupings of the same set of mentions in coreference clusters can lead to totally different meanings. Usually, to address this problem, many alternative hypotheses are generated and then scored. 
If multi-layer Transformer models indeed recover a form of an NLP pipeline,  their architecture forces the fusion of alternatives at every processing stage (after every layer). 
There is just no storage for alternative hypotheses.

In this work, we investigate the effects of allowing the Transformer encoder to preserve and explore alternative hypotheses, which can then be evaluated and combined at the end of the encoding process.  To that end, we propose an architecture we term \textit{Multi-stream Transformer} (see Figure \ref{fig:multistream_diag}), 
where only the first and the last layers compute a joint representation.  In the middle layers, the encoder is split into two or more encoder streams, where computation is not merged between the separate encoder layers.  
%
%
A given encoder layer in each encoder "stream" of the multi-stream part of the Transformer performs computation over the full input.  However, different encoder streams do not have access to, nor perform any computation over each other's representations. Results of parallel encoder stream computations are combined in the final layer of the encoder.

This can be thought of as allowing the model to retain multiple hypotheses about input representation, which can then be combined at the end, following the same general idea as stacking ensemble models leveraging multiple hypotheses.  As a \textit{zero-level hypothesis}, we use a skip connection from the first joint layer to the last joint layer, allowing the model to factor in the initial jointly computed representation, and at the same time improving learnability. 

Using one of the standard machine translation benchmarks, we show that \textit{Multi-stream Transformer} outperforms its vanilla equivalent "linear stream" Transformer.
We also show that adding a skip connection substantially improves performance of the vanilla Transformer baseline and has a comparable effect on the Multi-stream Transformer model.  To our knowledge, this is the first study that examines the effects of splitting the computation in Transformer models into multiple parallel streams.








\section{Model Architecture} 


As a starting point, we use vanilla Transformer \cite{vaswani2017attention}. We modify its encoder portion by splitting it into three segments: the input layer $L_{in}$, the body of the encoder with multiple \textit{streams}, and the output layer $L_{out}$. We use $S_i$ to denote $i$-th stream with output $Z_i$.

\begin{figure}
    \centering
    \includegraphics[width=\columnwidth]{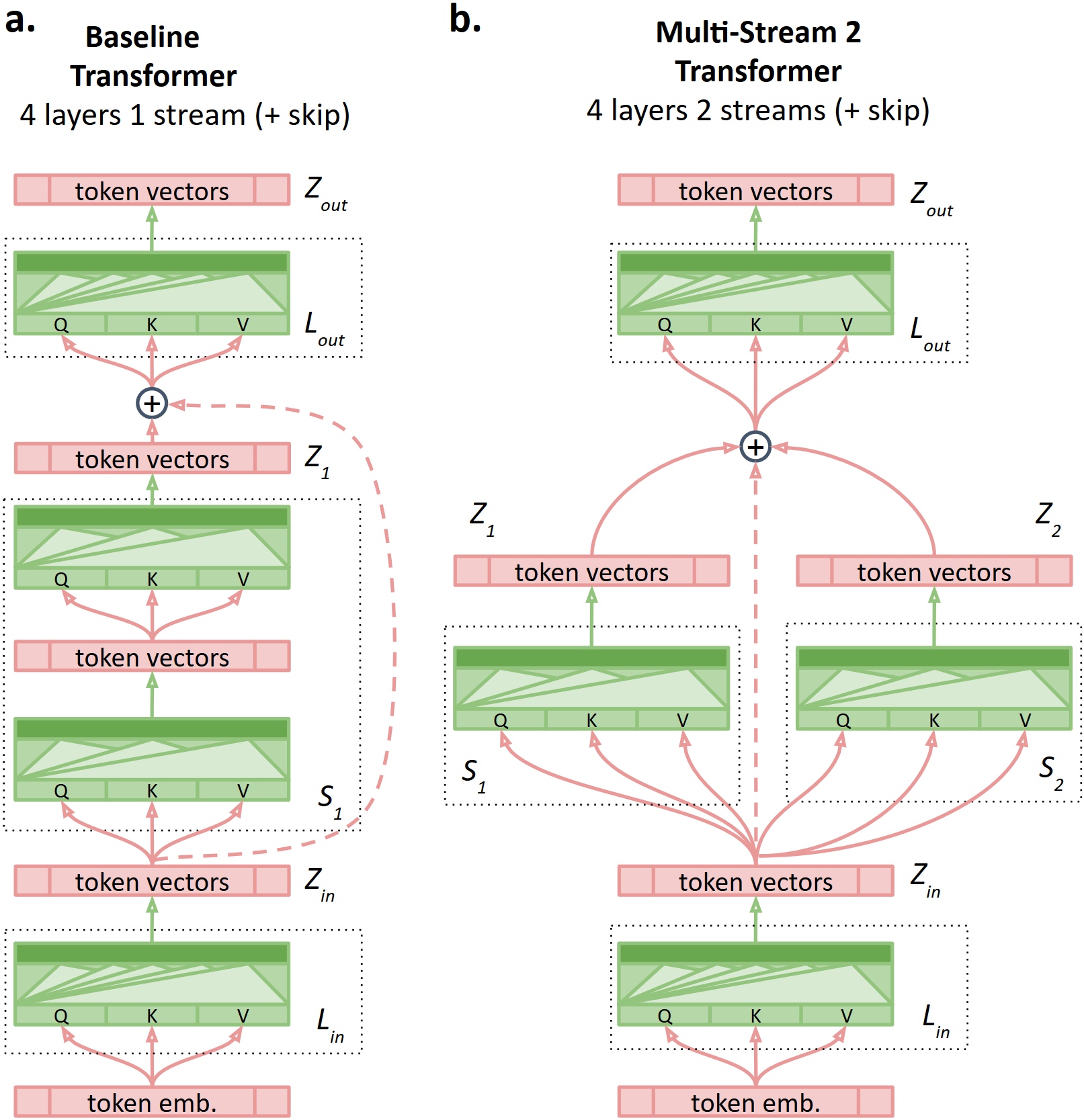}
    \caption{\textbf{Multi-Stream Transformer Encoder.} \\ 
    \textbf{a. }Transformer encoder is a sequence of layers, we use this architecture as a baseline. We differentiate input transformer layer $L_{in}$, \textit{streams}  $S_i$  and output transformer layer $L_{out}$ in the encoder. Baseline "linear stream" model has only one stream  $S_1$.  Skip connection is created if output of the input stream $Z_{in}$ is added to the output $Z_1$ of the linear stream. \textbf{b.} Multi-Stream encoder with two streams $S_{1}$ and $S_{2}$. Skip connection is shown as a dashed line.}
    \label{fig:multistream_diag}
\end{figure}

Figure \ref{fig:multistream_diag} shows a baseline "linear stream" encoder (\ref{fig:multistream_diag}a) and a multi-stream encoder with two parallel streams containing one layer each (\ref{fig:multistream_diag}b). The skip connection wrapping the body of the encoder from the input layer to the output layer is shown as a dashed line in both cases.
Output for the multi-stream encoder with a skip connection is calculated as:

\begin{equation}
    Z_{out} = L_{out} \left( \sum_{1 \leq i \leq k}{S_i(Z_{in})} + Z_{in}\right),
\end{equation}
where $k$ is the number of parallel streams.
We will refer to the Multi-Stream Transformer that has in encoder $k$ parallel streams of $l$ layers each as Multi-Stream $k(l)$. Figure \ref{fig:multistream_diag}b shows Multi-Stream 2(1) model with the total of 4 encoder layers.  See Appendix A for an illustration of Multi-Stream 2(2) and Multi-Stream 4(1) architectures with 6 layers.  Note that for this study, we only consider Multi-Stream Transformers with identical number of layers in all streams.


\section{Experiments} 

\subsection{Model configurations}

\begin{figure*}
    \centering
    \includegraphics[width=\textwidth]{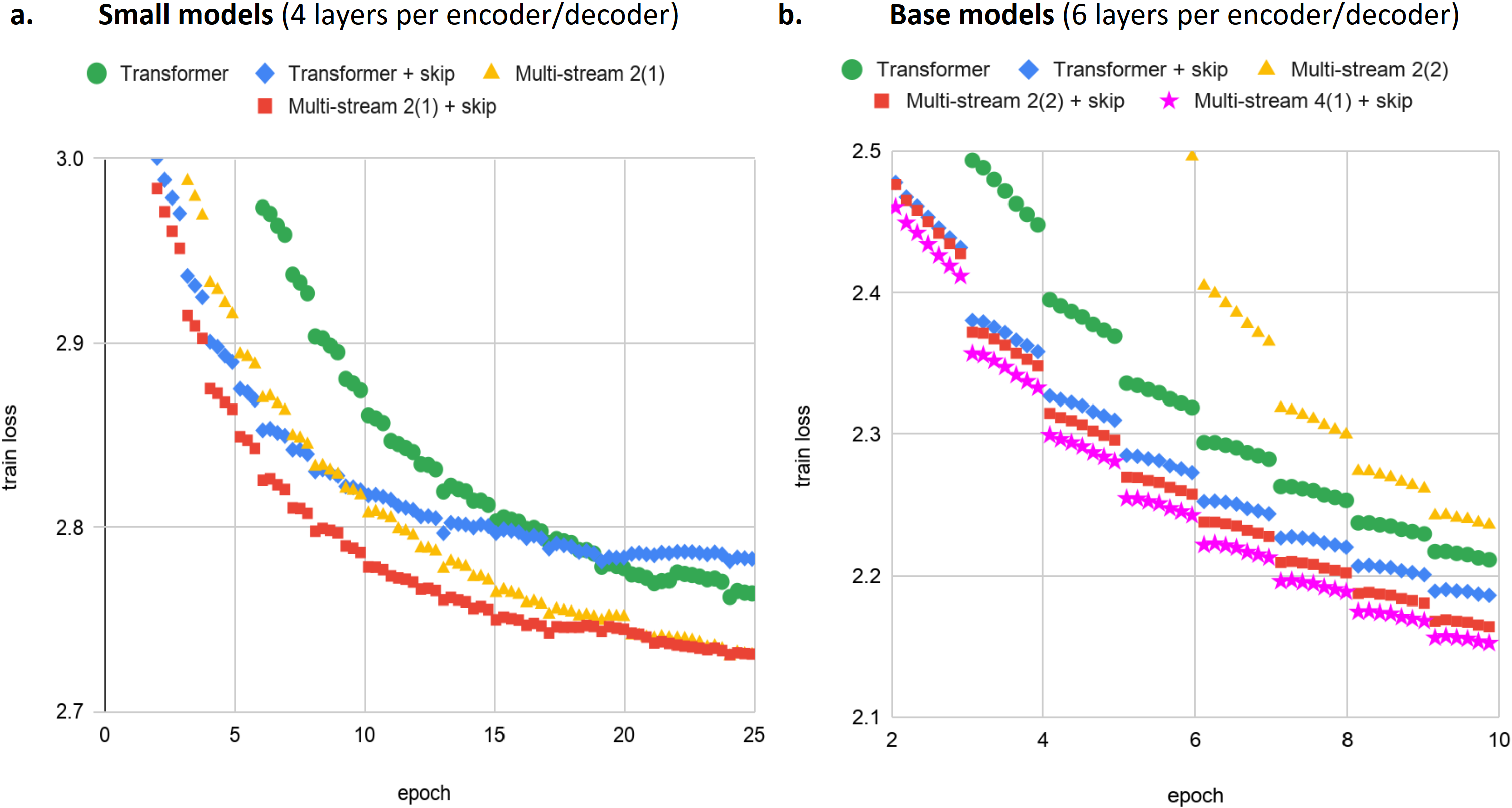}
    \caption{\textbf{Multi-stream Transformers with skip connection show better performance.} \textbf{a.} All multi-stream variations for the 4-layer model (see Figure \ref{fig:multistream_diag}) demonstrate faster convergence and better loss values after 20 epochs compared to vanilla Transformer. Adding only skip connection (Transformer-skip) accelerates learning on the first stage (10 epochs) but makes no difference after. \textbf{b.} Performance of the 6-layer models is significantly improved by presence of the skip connection. A model with two streams consisting of two layers each (Multi-stream 2(2)) shows the worst results. All data points presented are average of 3 runs for each model.} 
    \label{fig:train_loss}
\end{figure*}


We studied multi-stream models of two sizes: models with four encoder layers and with six encoder layers.  In both cases, we used a regular "linear stream" decoder with the same number of layers. Models with 4 layers (Figure \ref{fig:multistream_diag}b.) include Multi-Stream 2(1) and Multi-Stream 2(1) +skip, in which a skip connection is added. Models with 6 layers (Appendix A) include Multi-stream 2(2) with two layers per stream, Multi-stream 2(2) +skip, and Multi-stream 4(1) +skip. The last model has four single-layer streams and skip connection.
%
Baseline model in each case is a vanilla Transformer that has the same number of layers arranged in a standard linear stream.
Due to the computational constraints, we train 4-layer models for 20 epochs and 6-layer models for 10 epochs, using default parameters (see Appendix B).

We used a standard machine translation benchmark WMT-14 DE-EN \cite{bojar-EtAl:2014:W14-33} in our experiments. 

\subsection{Results}

Table \ref{table:bleu} presents averaged BLEU scores after the training for the studied models. Multi-stream architectures with skip connection significantly outperforms Transformer as well as Transformer with skip connection. We had not extensively tuned training parameters so the BLEU scores of baseline as well as multi-stream models of the 6-layer size were in the range of 24-25.5. This is a modest but solid performance given that recent results reported in the literature mainly vary between 20 and 30 of BLEU score (see for example \cite{kasai2020parallel}). 

Training loss for each model is shown in Figure \ref{fig:train_loss}. For both 4-layer and 6-layer sizes multi-stream architectures  demonstrate advantage over the baseline. Although average learning curve for  Multi-stream 2(2) converge slower compared to vanilla Transformer, closer examination of individual runs shows that 2 of them similar to the baseline and the one run is outlier. Experiments with 4-layer models (Figure \ref{fig:train_loss}a) also shows that training instability arise after 10-15 epochs of training especially for baseline Transformer and its skip connection modification. These results suggest that adding skip connection accelerates training but in the case of the baseline linear stream Transformer leads to inferior performance. However, note that multi-stream models with skip connections consistently converge faster and to a lower loss value for both 4- and 6-layer settings. 



\subsection{Attention patterns}

In order to better understand the differences in processing that occurs in different model configurations, we examined attention patterns for 6-layer Baseline, Baseline +skip and Multi-stream 4(1) +skip. Common pattern types are vertical, diagonal, diagonal shift forward, diagonal shift backward, soft diagonal and heterogeneous (see Figure 2 and 3 in the Appendix C for details).  Diagonal patterns effectively copy information from the closest neighbours or the token itself. Heterogeneous patterns aggregate information over the whole sequence, and it has been argued that only heterogeneous attention heads may track linguistic structure \cite{KovalevaRomanovEtAl_2019_Revealing_Dark_Secrets_of_BERT}.
Interestingly, some layers have the tendency for the majority of heads to have the same type of attention pattern suggesting that different layers may indeed perform specialized processing.

We note that self-attention heads in the first layer of the encoder are similar for all models. 
However, patterns observed in layers 2--5 of the encoder differ across models.  
For the Baseline, heterogeneous patterns disappear and diagonal patterns become dominant in the later layers.  Vertical patterns dominate the Baseline skip model for all intermediate layers.
In Multi-stream 4(1) +skip models, on the other hand, all patterns types seem to occur equally often. Visual examination of attention patterns in different streams in this model shows that attention patterns indeed vary between different streams, suggesting that different streams may be processing alternative interpretations of the input.

%


\begin{table}[]
\centering
\begin{tabular}{ll}
\hline \hline
\multicolumn{2}{c}{\textbf{4 layers per encoder/decoder}}   \\
\multicolumn{2}{c}{(20 epochs)} \\ \hline
Transformer                & 18.99            \\
Transformer +skip         & 18.76            \\
Multi-stream 2(1)              & 19.13            \\
Multi-stream 2(1) +skip  & \textbf{19.46}   \\ \hline
\multicolumn{2}{c}{\textbf{6 layers per encoder/decoder}}   \\
\multicolumn{2}{c}{(10 epochs)} \\ \hline
Transformer          & 24.65             \\
Transformer +skip      & 25.49            \\
Multi-stream 2(2)       & 23.90                 \\
Multi-stream 2(2) +skip  & 25.61            \\
Multi-stream 4(1) +skip  & \textbf{25.63}   \\ 
\hline \hline
\end{tabular}
\caption{\textbf{Performance of Multi-Stream Transformer models on WMT-14 DE-EN translation task.} Values represent an average of BLEU 4 scores for 3 runs of every model. }
\label{table:bleu}
\end{table}

\section{Discussion}

Our experiments suggest that Multi-Stream Transformer has a competitive advantage over the corresponding linear stream baseline, with the advantage becoming more pronounced in the presence of skip connection between the first and the last layers of the Transformer. 

Following several recent studies that analyse different types of attention patterns observed in Transformer architectures~\cite{KovalevaRomanovEtAl_2019_Revealing_Dark_Secrets_of_BERT,clark2019does}, we examined the attention patterns obtained when the decoder attention heads are attending to the last layer of the encoder.  
\newcite{KovalevaRomanovEtAl_2019_Revealing_Dark_Secrets_of_BERT} proposed to classify self-attention patterns into five categories, surmising that only "heterogeneous" patterns which spread attention across all input tokens might track non-trivial information about linguistic structure.  
Among other classes, "vertical" patterns characterized the situation when the attention was directed solely at a particular token (usually, SEP, CLS, or punctuation).  
In the BERT model \cite{devlin2019bert} fine-tuned for GLUE \cite{WangSinghEtAl_2018_GLUE_A_Multi-Task_Benchmark_and_Analysis_Platform_for_Natural_Language_Understanding},  they were estimated to account for about 30\% of all attention patterns \cite{KovalevaRomanovEtAl_2019_Revealing_Dark_Secrets_of_BERT}.

Recently, \newcite{KobayashiKuribayashiEtAl_2020_Attention_Module_is_Not_Only_Weight_Analyzing_Transformers_with_Vector_Norms} re-examined vertical patterns looking at the norms of the attention-weighted input vectors, rather than just the attention weights. 
They found that when "vertical" attention heads pay attention to \texttt{SEP}s and other special tokens, attention weights and input vector norms are inversely correlated.
That is, "vertical" self-attention patterns function as a "pass-through" operation that collects no information from the input tokens. Self-attention computation is essentially ignored, and the skip connection passes through the information computed in the previous layer.

Examining decoder attention patterns over encoder output in our models, we observed that linear stream baseline Transformer tends to have substantially more of such vertical patterns, while the Multi-stream Transformer tends to be dominated by "heterogeneous" patterns (see Figure 3 in Appendix C). 
%
%
If the hypothesis that heterogeneous self-attention patterns may encode meaningful linguistic information is correct, this may be seen as supporting the hypothesis that the encoder in multi-stream Transformer is delaying the resolution of ambiguities in the input.



\section{Conclusion}

This study has shown that Multi-Stream Transformer models that take advantage of separate parallel streams of computation are able to outperform equivalent linear stream Transformers.  We also presented some evidence in support of the notion that parallel streams compute alternative hypothesis which are evaluated when they are merged at the final layer of the encoder.

\bibliographystyle{acl_natbib}
\bibliography{anthology,emnlp2020}


\begin{figure*}
    \centering
    \includegraphics[width=\textwidth]{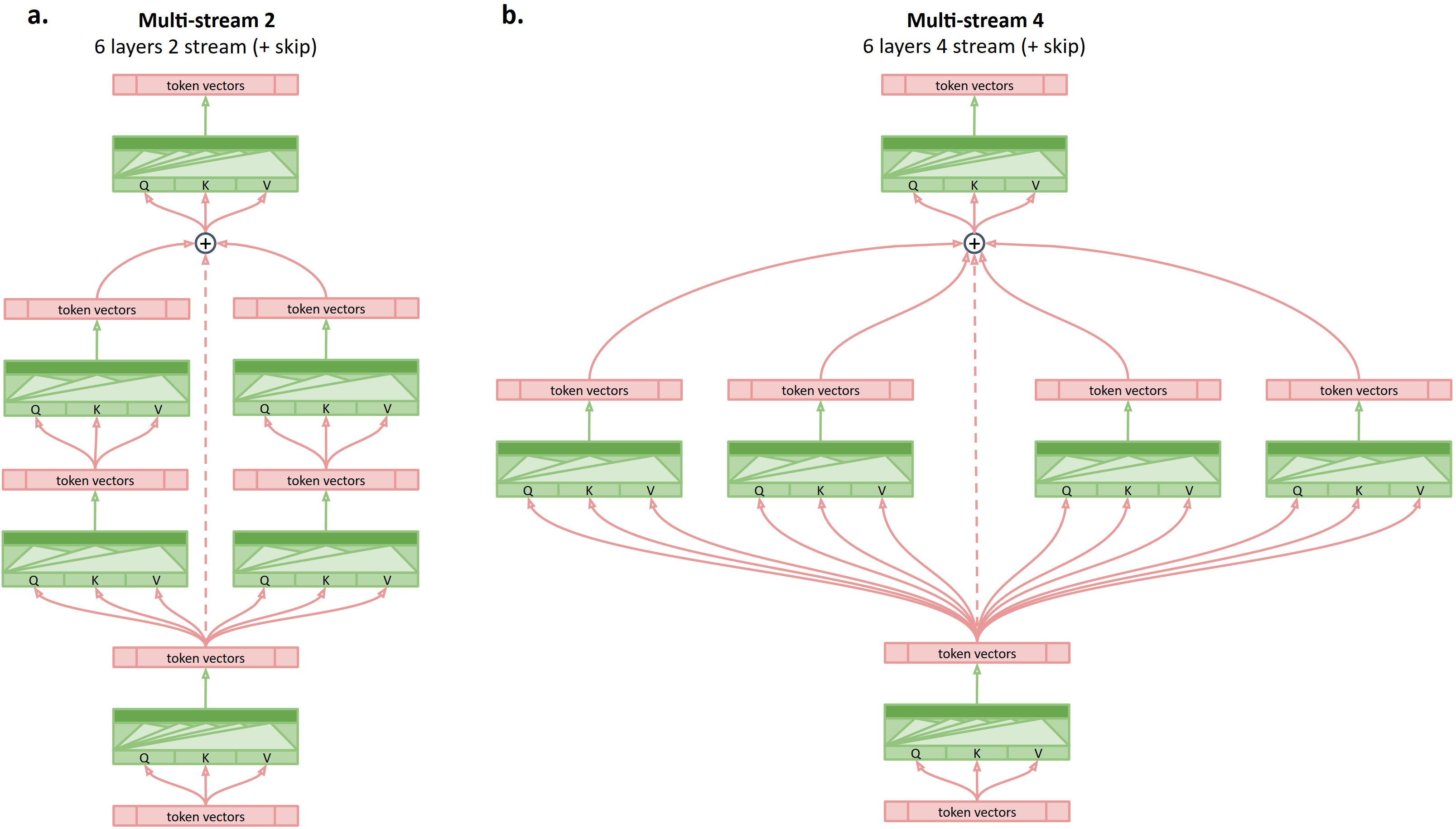}
    \caption{\textbf{Multi-Stream Transformer Encoder with 6 layers.} \textbf{a. }Multi-stream 2(2) encoder consists of two streams of depth 2. \textbf{b.} Multi-stream 4(1) encoder has 4 one layer streams.}
    \label{fig:multistream_diag_6l}
\end{figure*}

\newpage

\appendix

\section{Multi-stream architectures for 6 encoder layers}
\label{sec:app_6streams_figure}

Figure \ref{fig:multistream_diag_6l} shows Multi-Stream Transformer architectures with 6 encoder layers.

\section{Training details}
\label{sec:app_train_det}

To implement multi-stream Transformer, we extended the code from the dedicated TensorFlow tutorial repository\footnote{\url{https://github.com/tensorflow/docs/blob/master/site/en/tutorials/text/transformer.ipynb}}.


Models of both sizes used default settings. The 4 layer setup had parameters $d_{model}=128, d_{ff}=512, h = 8, P_{drop} = 0.1, warmup_{steps} = 4000$, and the 6 layer setup had $d_{model}=512, d_{ff}=2048, h = 8, P_{drop} = 0.1, warmup_{steps} = 32000$. For all experiments batch size was 64.

\newpage

We used a standard machine translation benchmark WMT-14 DE-EN \cite{bojar-EtAl:2014:W14-33} in our experiments. One epoch of training covered the full training set of 4.5M sentence pairs. Dataset was tokenized with TFDS subword text encoder\footnote{\url{https://www.tensorflow.org/datasets/api_docs/python/tfds/features/text/SubwordTextEncoder}} and dictionary of 32K per language.


\section{Visualization of attention patterns.}
\label{sec:appendix}

\begin{figure*}
    \centering
    \includegraphics[width=\textwidth]{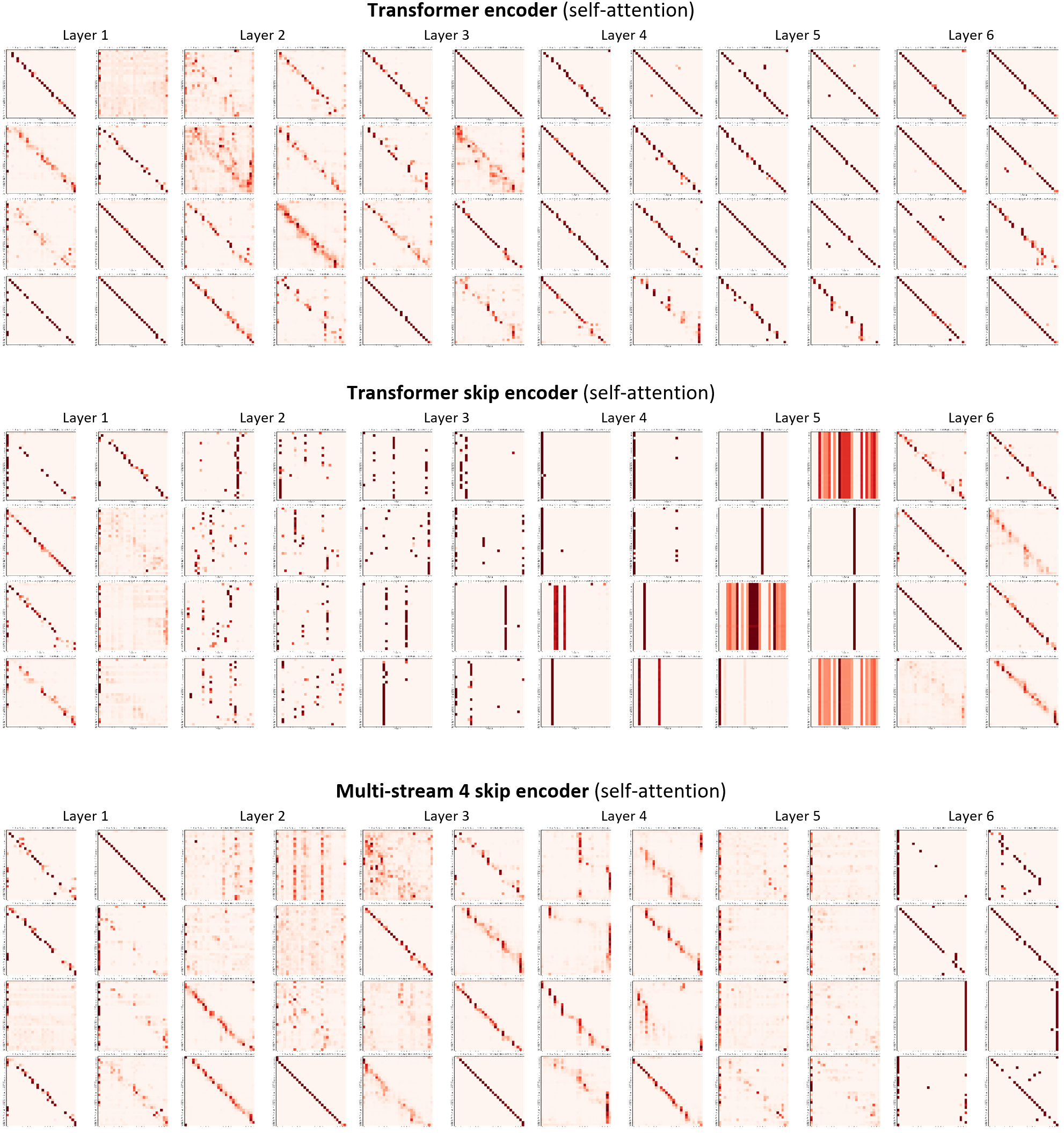}
    \caption{Visualization of the baseline Transformer, baseline Transformer skip and Multi-Stream 4(1) +skip encoder self-attention patterns.} 
    \label{fig:att_enc}
\end{figure*}

\begin{figure*}
    \centering
    \includegraphics[width=\textwidth]{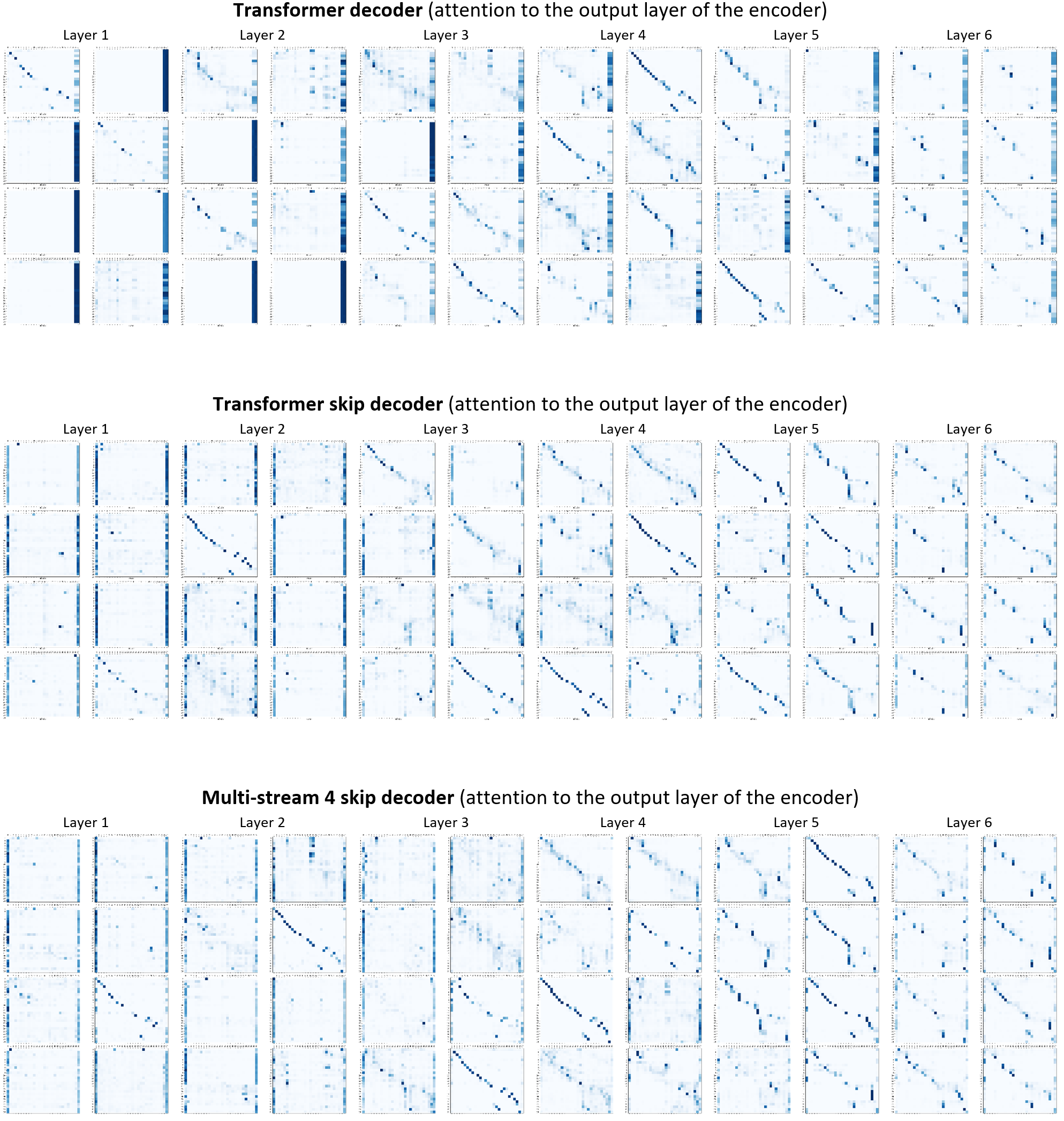}
    \caption{Transformer, Transformer skip and Multi-stream 4(1) +skip attention patterns from the different decoder layers to encoder output layer.} 
    \label{fig:att_dec}
\end{figure*}

Figures \ref{fig:att_enc} and \ref{fig:att_dec} show (a) encoder self-attention patterns and (b) decoder attention to encoder output layer for baseline linear stream Transformer with and without skip connection, and for the Multi-Stream 4(1) architecture.
%

\end{document}